\pdfoutput=1
\documentclass[11pt]{article}

\usepackage[table,xcdraw]{xcolor}
\usepackage{acl}

\usepackage{times}
\usepackage{latexsym}
\usepackage{booktabs}
\usepackage{multirow}

\usepackage[T1]{fontenc}
\usepackage[utf8]{inputenc}
\DeclareUnicodeCharacter{307}{\^{}}

\usepackage{microtype}
\usepackage{url}
\usepackage{graphicx}

\usepackage{ stmaryrd }

\newcommand{\DatasetName}{LMGQS}
\newcommand{\DatasetNameSpace}{LMGQS }

\title{\DatasetName: A Large-scale Dataset for Query-focused Summarization}

\author{Ruochen Xu, Song Wang, Yang Liu, Shuohang Wang, Yichong Xu, Dan Iter \AND Chenguang Zhu, Michael Zeng\\
Microsoft\\
  \texttt{\{ruox,sonwang,yaliu10,shuowa,yicxu,iterdan,chezhu,nzeng\}@microsoft.com} \\}

\begin{document}
\maketitle
\begin{abstract}
Query-focused summarization (QFS) aims to extract or generate a summary of an input document that directly answers or is relevant to a given query. The lack of large-scale datasets in the form of documents, queries, and summaries has hindered model development in this area. In contrast, multiple large-scale high-quality datasets for generic summarization exist.
We hypothesize that there is a hidden query for each summary sentence in a generic summarization annotation, and we utilize a large-scale pretrained language model to recover it. In this way, we convert four generic summarization benchmarks into a new QFS benchmark dataset, \DatasetName, which consists of over 1 million document-query-summary samples. We thoroughly investigate the properties of our proposed dataset and establish baselines with state-of-the-art summarization models. By fine-tuning a language model on \DatasetName, we achieve state-of-the-art zero-shot and supervised performance on multiple existing QFS benchmarks, demonstrating the high quality and diversity of \DatasetName.
\end{abstract}

\section{Introduction}
The field of generic summarization \citep{see-etal-2017-get,gehrmann-etal-2018-bottom,liu-lapata-2019-text} has made significant progress in recent years, thanks to the development of generative deep neural models \citep{NIPS2014_a14ac55a,NIPS2017_3f5ee243} and the availability of large-scale training data \citep{nallapati-etal-2016-abstractive,narayan-etal-2018-dont,zhu-etal-2021-mediasum}. However, query-focused summarization (QFS) presents a significant challenge due to the lack of data. Most of the available QFS corpora \citep{dang-2006-duc,dang2006duc,nema-etal-2017-diversity,baumel2016topic,zhong-etal-2021-qmsum} contain only a few thousand documents or less, which is insufficient for training a robust neural model.

\begin{figure}[!ht]
\centering
\resizebox{0.95\linewidth}{!}{
\includegraphics{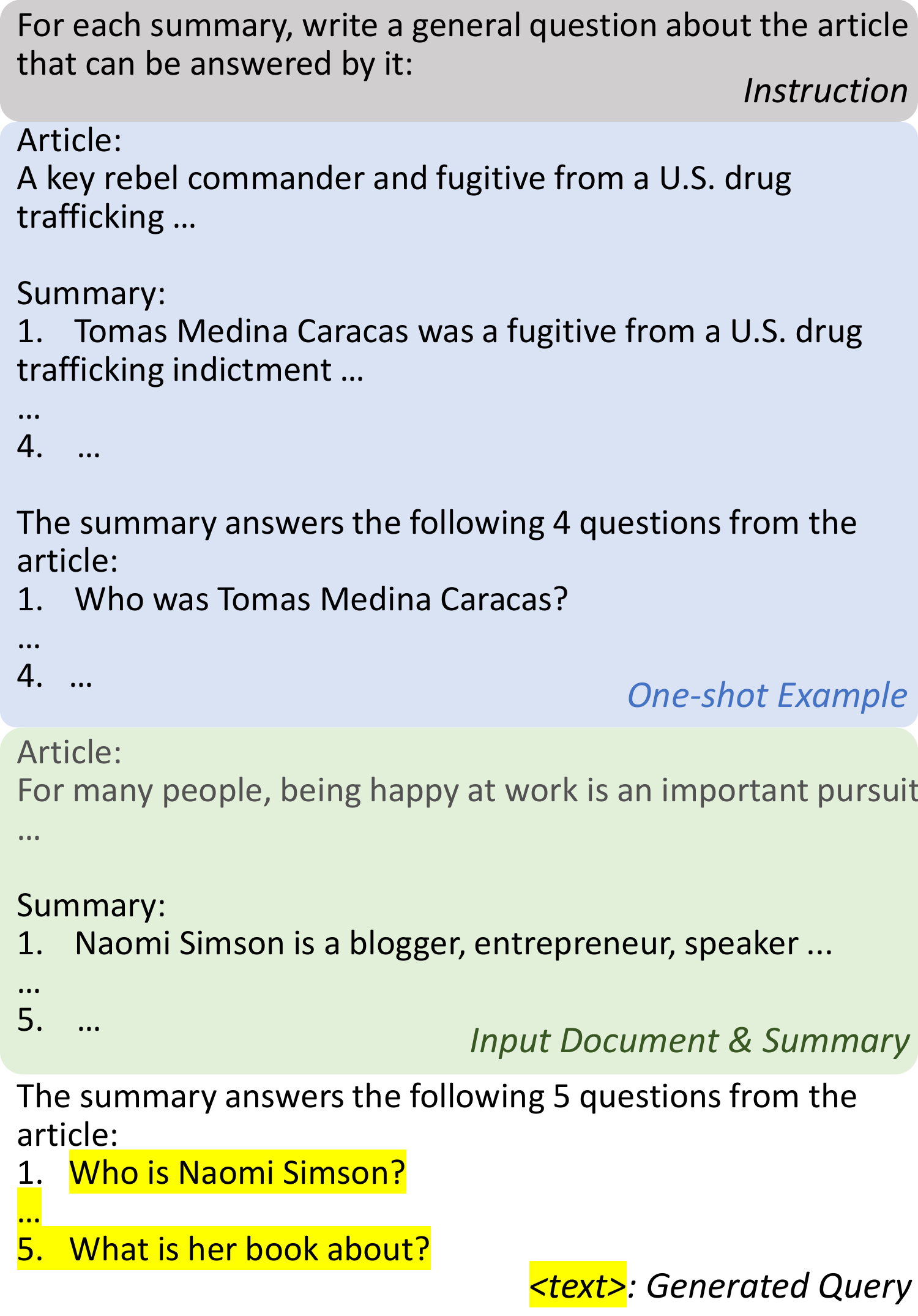}
}
\caption{
% \textcolor{blue}
% {cz:the legend is too small. suggest putting section name directly by the rectangle above, like One-Shot Example, ...}
Example prompt for query generation. The top part is the instruction, followed by the one-shot example consisting of document, summary, and query. The query for the input document (highlighted in yellow) and summary is generated by InstructGPT.}
\label{fig:prompt exmpl}
\end{figure}

% \usepackage{multirow}
% \usepackage{graphicx}
\begin{table*}[ht]
\centering
\resizebox{\textwidth}{!}{%
\begin{tabular}{c|lllllllllllllll|ll}
\hline
\multicolumn{1}{l|}{\begin{tabular}[c]{@{}l@{}}Prompt \\ Query Type\end{tabular}} &
  Dataset &
  \begin{tabular}[c]{@{}l@{}}do\\ does\\ did\end{tabular} &
  \begin{tabular}[c]{@{}l@{}}is/are\\ was\\ were\end{tabular} &
  \begin{tabular}[c]{@{}l@{}}can\\ could\end{tabular} &
  \begin{tabular}[c]{@{}l@{}}will\\ would\end{tabular} &
  \begin{tabular}[c]{@{}l@{}}have\\ has\\ had\end{tabular} &
  what &
  when &
  where &
  \begin{tabular}[c]{@{}l@{}}who\\ whom\end{tabular} &
  which &
  whose &
  why &
  how &
  other &
  \begin{tabular}[c]{@{}l@{}}Yes/no\\ queries\end{tabular} &
  \begin{tabular}[c]{@{}l@{}}Wh-\\ queries\end{tabular} \\ \hline
 &
  CNN/DM &
  \cellcolor[HTML]{F9B0B2}0.08 &
  \cellcolor[HTML]{FBF7FA}0.16 &
  \cellcolor[HTML]{F87173}0.01 &
  \cellcolor[HTML]{F87A7C}0.02 &
  \cellcolor[HTML]{F87A7C}0.02 &
  \cellcolor[HTML]{91B1DA}52.36 &
  \cellcolor[HTML]{F2F5FC}5.36 &
  \cellcolor[HTML]{F4F7FD}4.29 &
  \cellcolor[HTML]{E0E8F5}14.21 &
  \cellcolor[HTML]{FCFCFF}0.2 &
  \cellcolor[HTML]{F8696B}0 &
  \cellcolor[HTML]{F5F7FD}3.98 &
  \cellcolor[HTML]{D5E1F2}19.21 &
  \cellcolor[HTML]{FAC2C4}0.1 &
  \cellcolor[HTML]{F8696B}0.29 &
  \cellcolor[HTML]{5A8AC6}99.61 \\
 &
  XSUM &
  \cellcolor[HTML]{F99EA0}0.06 &
  \cellcolor[HTML]{FAB9BB}0.09 &
  \cellcolor[HTML]{F87173}0.01 &
  \cellcolor[HTML]{FAB9BB}0.09 &
  \cellcolor[HTML]{F87173}0.01 &
  \cellcolor[HTML]{6894CB}72.54 &
  \cellcolor[HTML]{F9FAFE}2.07 &
  \cellcolor[HTML]{FBFBFF}0.98 &
  \cellcolor[HTML]{E9EFF9}9.66 &
  \cellcolor[HTML]{FBF7FA}0.16 &
  \cellcolor[HTML]{F8696B}0 &
  \cellcolor[HTML]{F8FAFE}2.15 &
  \cellcolor[HTML]{E4EBF7}12.01 &
  \cellcolor[HTML]{FCFCFF}0.17 &
  \cellcolor[HTML]{F8696B}0.26 &
  \cellcolor[HTML]{5B8BC7}99.57 \\
 &
  SAMSum &
  \cellcolor[HTML]{FCFCFF}0.49 &
  \cellcolor[HTML]{FCFCFF}0.19 &
  \cellcolor[HTML]{F88385}0.03 &
  \cellcolor[HTML]{FAD3D6}0.12 &
  \cellcolor[HTML]{F88C8E}0.04 &
  \cellcolor[HTML]{749DD0}66.43 &
  \cellcolor[HTML]{EEF2FA}7.12 &
  \cellcolor[HTML]{EFF3FB}6.58 &
  \cellcolor[HTML]{EDF2FA}7.6 &
  \cellcolor[HTML]{FAD3D6}0.12 &
  \cellcolor[HTML]{F8696B}0 &
  \cellcolor[HTML]{F7F9FE}2.79 &
  \cellcolor[HTML]{ECF1FA}8.29 &
  \cellcolor[HTML]{FCFCFF}0.19 &
  \cellcolor[HTML]{F86B6D}0.87 &
  \cellcolor[HTML]{5D8CC7}98.93 \\
\multirow{-4}{*}{Wh-queries} &
  DialogSum &
  \cellcolor[HTML]{FCFCFF}0.44 &
  \cellcolor[HTML]{FBDCDF}0.13 &
  \cellcolor[HTML]{F88C8E}0.04 &
  \cellcolor[HTML]{F87A7C}0.02 &
  \cellcolor[HTML]{F87A7C}0.02 &
  \cellcolor[HTML]{5A8AC6}78.88 &
  \cellcolor[HTML]{FAFBFF}1.4 &
  \cellcolor[HTML]{F7F9FE}2.82 &
  \cellcolor[HTML]{F5F7FD}3.8 &
  \cellcolor[HTML]{FAD3D6}0.12 &
  \cellcolor[HTML]{F8696B}0 &
  \cellcolor[HTML]{F5F7FD}3.88 &
  \cellcolor[HTML]{ECF1FA}8.35 &
  \cellcolor[HTML]{F9B0B2}0.08 &
  \cellcolor[HTML]{F86B6D}0.65 &
  \cellcolor[HTML]{5C8BC7}99.25 \\ \hline
 &
  CNN/DM &
  \cellcolor[HTML]{AAC2E2}40.33 &
  \cellcolor[HTML]{A4BEE0}43.06 &
  \cellcolor[HTML]{FAFBFF}1.54 &
  \cellcolor[HTML]{F1F4FB}5.79 &
  \cellcolor[HTML]{F4F6FC}4.42 &
  \cellcolor[HTML]{FAD3D6}0.12 &
  \cellcolor[HTML]{FCFCFF}0.17 &
  \cellcolor[HTML]{F87A7C}0.02 &
  \cellcolor[HTML]{FBEEF1}0.15 &
  \cellcolor[HTML]{F87173}0.01 &
  \cellcolor[HTML]{F8696B}0 &
  \cellcolor[HTML]{F87173}0.01 &
  \cellcolor[HTML]{F9B0B2}0.08 &
  \cellcolor[HTML]{F4F6FC}4.32 &
  \cellcolor[HTML]{6894CB}95.14 &
  \cellcolor[HTML]{F86A6C}0.56 \\
 &
  XSUM &
  \cellcolor[HTML]{C6D6EC}26.56 &
  \cellcolor[HTML]{A6C0E1}42.19 &
  \cellcolor[HTML]{FAFBFF}1.41 &
  \cellcolor[HTML]{E4EBF7}12.2 &
  \cellcolor[HTML]{ECF1FA}8.07 &
  \cellcolor[HTML]{FCFCFF}0.17 &
  \cellcolor[HTML]{FAB9BB}0.09 &
  \cellcolor[HTML]{F87173}0.01 &
  \cellcolor[HTML]{FBDCDF}0.13 &
  \cellcolor[HTML]{F8696B}0 &
  \cellcolor[HTML]{F8696B}0 &
  \cellcolor[HTML]{F8696B}0 &
  \cellcolor[HTML]{F99597}0.05 &
  \cellcolor[HTML]{EAF0F9}9.12 &
  \cellcolor[HTML]{769ED0}90.43 &
  \cellcolor[HTML]{F86A6C}0.45 \\
 &
  SAMSum &
  \cellcolor[HTML]{AFC6E4}37.96 &
  \cellcolor[HTML]{B1C7E5}36.78 &
  \cellcolor[HTML]{FBFCFF}0.79 &
  \cellcolor[HTML]{D8E3F3}17.88 &
  \cellcolor[HTML]{F8F9FE}2.52 &
  \cellcolor[HTML]{F88385}0.03 &
  \cellcolor[HTML]{F88385}0.03 &
  \cellcolor[HTML]{F87173}0.01 &
  \cellcolor[HTML]{F8696B}0 &
  \cellcolor[HTML]{F8696B}0 &
  \cellcolor[HTML]{F8696B}0 &
  \cellcolor[HTML]{F8696B}0 &
  \cellcolor[HTML]{F87173}0.01 &
  \cellcolor[HTML]{F5F7FD}3.99 &
  \cellcolor[HTML]{6692CA}95.93 &
  \cellcolor[HTML]{F8696B}0.08 \\
\multirow{-4}{*}{Yes/no queries} &
  DialogSum &
  \cellcolor[HTML]{82A6D4}59.68 &
  \cellcolor[HTML]{C4D4EB}27.84 &
  \cellcolor[HTML]{FCFCFF}0.56 &
  \cellcolor[HTML]{F3F6FC}4.9 &
  \cellcolor[HTML]{FBFBFF}1.11 &
  \cellcolor[HTML]{F87173}0.01 &
  \cellcolor[HTML]{F8696B}0 &
  \cellcolor[HTML]{F8696B}0 &
  \cellcolor[HTML]{F8696B}0 &
  \cellcolor[HTML]{F8696B}0 &
  \cellcolor[HTML]{F8696B}0 &
  \cellcolor[HTML]{F8696B}0 &
  \cellcolor[HTML]{F8696B}0 &
  \cellcolor[HTML]{F1F4FB}5.9 &
  \cellcolor[HTML]{6B96CC}94.09 &
  \cellcolor[HTML]{F8696B}0.01 \\ \hline
\end{tabular}%
}
\caption{Breakdown of query types on QuerySum. The upper part of the table shows the query type percentage when using wh-queries in the one-shot prompt example. The lower part shows the percentage when prompting with yes/no queries. Color of blue means a higher percentage while red means a lower one.}
\label{tab:query_type}
\end{table*}

We propose a Language Model Generated Query-Focused Summarization Dataset (\DatasetName) to address the lack of a large-scale QFS dataset. Human annotation for QFS typically involves generating suitable queries and then writing corresponding summaries, which is both time-consuming and expensive. Furthermore, it may necessitate a meticulous definition of the query scheme based on the domain of documents \citep{zhong-etal-2021-qmsum}. We hypothesize that, for a pair of document and summary in a generic summarization dataset, hidden queries exist that represent the information needs associated with the summary. Therefore, to efficiently scale annotation, we prompt the large-scale language model InstructGPT \citep{ouyang2022training} with documents and summaries from four generic summarization datasets to extract the hidden queries.
% \cz{any filtering process is applied?}
% no filtering is applied
This approach results in the \DatasetNameSpace dataset, which contains over 1.1 million triplets of document, query, and summary, encompassing a wide range of document and question types.
%Thus, we prompt large-utilize the strong few-shot capability of the large-scale language model \cite{NEURIPS2020_1457c0d6,zhao2021calibrate} 
%and include the document and the summary in the prompt. The language model completes the prompt with the missing query, 

To investigate the utility of our proposed \DatasetNameSpace, we finetune a pretrained language model on it. The model accepts the concatenation of the original document and generated query as input and is trained to produce the original summary. We then compare the finetuned model with various query-focused summarization models on several existing QFS benchmarks that have no overlap with \DatasetNameSpace under the zero-shot setting. 
%which means our model has no access to the QFS annotation on the downstream task and has to rely on the generalization and coverage of \DatasetNameSpace to generate sensible query-dependent summaries. 
Empirical results demonstrate that the model finetuned on \DatasetNameSpace achieves promising performance on both single-document and multi-document QFS benchmarks, surpassing strong baselines. Similarly, when utilizing \DatasetNameSpace for pre-finetuning, the model achieves state-of-the-art performance in the supervised setting.

In summary, our contributions are three-fold: (1) We introduce a novel framework for constructing a QFS dataset by converting existing generic summarization datasets using language models as annotators. (2) We present \DatasetNameSpace, a large-scale QFS benchmark, to foster future research on QFS\footnote{Dataset will be released after the anonymous period}. (3) The model finetuned on \DatasetNameSpace exhibits robust generalization capability and achieves remarkable zero-shot and supervised performance on other unseen QFS test sets.

% \section{Related Work}
\section{Dataset Creation}

We choose 4 generic datasets to build \DatasetName: CNN/DailyMail \cite{nallapati-etal-2016-abstractive}, XSUM \cite{narayan-etal-2018-dont}, SAMSum \cite{gliwa-etal-2019-samsum}, and DialogSum \cite{chen-etal-2021-dialogsum}. Among them, CNN/DailyMail and XSUM are news summarization datasets, where both the documents and summaries are in formal written English. SAMSum and DialogSum are two recently proposed dialogue summarization datasets, whose inputs are the transcripts of multi-speaker conversations.

\subsection{Prompt-based Query Generation}
\label{subsec:query_gen}
Given a document and its corresponding summary, we take advantage of the robust few-shot capabilities of InstructGPT \citep{ouyang2022training} to generate a query that encapsulates the information required by the annotator when crafting the summary. More specifically, we construct a prompt for each document-summary pair and input it into the InstructGPT model to generate the query by completing the prompt. An example prompt is illustrated in Figure \ref{fig:prompt exmpl}.
Since InstructGPT excels at adhering to human-readable instructions and can even generalize to unseen instructions \cite{ouyang2022training}, we begin our prompt with a clear directive for the query generation task. Following the instruction, we incorporate a one-shot example of the task into the prompt, which includes a human-written query derived from a document-summary pair. We set the number of examples to be $1$ based on a balance between effectiveness and efficiency: during our preliminary exploration, we noticed more failure cases for zero-shot query generation, while incorporating additional examples in the prompt would increase both the time and cost of generation.

% \usepackage{booktabs}
% \usepackage{multirow}
% \usepackage{graphicx}
\begin{table*}[th]
\centering
\resizebox{0.98\textwidth}{!}{%
\begin{tabular}{@{}c|lllllllllll@{}}
\toprule
\multicolumn{1}{l|}{Prompt   Query Type} &
  Dataset &
  Count &
  \begin{tabular}[c]{@{}l@{}}Len\\ (doc)\end{tabular} &
  \begin{tabular}[c]{@{}l@{}}Len\\ (query)\end{tabular} &
  \begin{tabular}[c]{@{}l@{}}Len\\ (sum)\end{tabular} &
  \begin{tabular}[c]{@{}l@{}}NTP\\ (sum, doc)\end{tabular} &
  \begin{tabular}[c]{@{}l@{}}NTP\\ (query,   doc)\end{tabular} &
  \begin{tabular}[c]{@{}l@{}}NTP\\ (doc, sum)\end{tabular} &
  \begin{tabular}[c]{@{}l@{}}NTP\\ (doc, query)\end{tabular} &
  \begin{tabular}[c]{@{}l@{}}NTP\\ (query, sum)\end{tabular} &
  \begin{tabular}[c]{@{}l@{}}NTP\\ (sum, query)\end{tabular} \\ \midrule
\multirow{4}{*}{Wh-queries}     & CNN/DM    & 311938 & 677.7 & 29.9 & 52.0 & 21.5 & 36.7 & 87.34 & 94.74 & 44.6 & 70.3 \\
                                & XSUM      & 226287 & 368.0 & 10.0 & 22.1 & 42.2 & 43.1 & 92.37 & 96.45 & 45.9 & 73.4 \\
                                & SAMSum    & 16368  & 92.9  & 16.2 & 22.3 & 41.5 & 56.5 & 76.51 & 87.89 & 43.5 & 59.2 \\
                                & DialogSum & 14460  & 130.7 & 15.1 & 24.2 & 36.6 & 48.0 & 82.7  & 90.97 & 47.0 & 65.9 \\ \midrule
\multirow{4}{*}{Yes/no queries} & CNN/DM    & 311920 & 677.7 & 33.7 & 52.0 & 21.5 & 27.4 & 87.34 & 92.6  & 20.7 & 48.2 \\
                                & XSUM      & 226276 & 368.0 & 10.7 & 22.1 & 42.2 & 40.4 & 92.37 & 95.91 & 21.4 & 57.8 \\
                                & SAMSum    & 16368  & 92.9  & 17.5 & 22.3 & 41.5 & 42.9 & 76.51 & 81.71 & 19.0 & 34.1 \\
                                & DialogSum & 14460  & 130.7 & 16.4 & 24.2 & 36.6 & 37.9 & 82.7  & 87.91 & 27.8 & 47.9 \\ \bottomrule
\end{tabular}%
}
\caption{Statistics about \DatasetName. "Len" stands for length and Len(string) is the length of the string. "NTP" stands for novel token percentage. NTP(string1, string2) computes the percentage of tokens in string1 that are not present in string2.}
\label{tab:dataset_stat}
\end{table*}

% \cz{better to be number of queries?}
In the one-shot example, we restrict the number of queries to be equivalent to the number of summaries. In other words, there exists a one-to-one correspondence between the sentences in the summary and the query. This constraint is imposed by appending prefix indices and appending newline characters for each summary/query sentence, as illustrated in Figure~\ref{fig:prompt exmpl}.

Due to the domain difference between news and dialogue summarization, we choose different one-shot examples for the two domains. The queries for the two example pairs were annotated by the authors of this paper. 

\subsection{Prompt Query Types}
Given a document and a summary sentence, multiple valid queries can be formulated. For instance, consider the summary sentence: \textit{She has released a book to encourage people to find their passion at work}. One possible query is: \textit{What is her book about?} Alternatively, another valid query could be: \textit{Has she released a book?} To address this variety, we utilize two sets of annotated queries: yes/no queries and wh-queries. Yes/no queries correspond to questions that can be answered with a simple "yes" or "no". However, in the context of QFS, the summary (i.e., the answer to the yes/no query) is never a mere "yes" or "no". For example, for a yes/no query like \textit{Is he still alive?}, we expect the answer to be: \textit{He was killed in an attack on a guerrilla encampment} rather than a simple \textit{no}.
% Wh-queries pertain to questions that begin with words such as who, what, where, when, why, how, how many, etc. 
Detailed annotated queries are presented in Table \ref{tab:prompt_example}.

The type of queries in a one-shot prompt significantly influences the generated queries. We provide a breakdown of query types in Table \ref{tab:query_type}. It is evident that when the prompt includes only wh-queries, over 99\% of the generated queries are also wh-queries, with the most frequent ones beginning with "What". The same pattern applies when the prompt contains only yes/no queries. The most common queries generated by InstructGPT typically start with "do/does/did" or "is/are/was/were".

\subsection{Statistics of LMGQS}
% \cz{Using the above prompting method, we collected XXX document-query-summary triples covering YYY different query tpes.} 
Using the aforementioned prompting method, we collected 1138077
 document-query-summary triples covering 13 different query types.
Detailed statistics of the generated \DatasetNameSpace dataset are shown in Table \ref{tab:dataset_stat}.
First, the length of the generated queries has a strong Pearson's correlation (0.95) with the length of summaries, which is expected due to our one-to-one mapping between the summary and query sentences.
Second, the length of queries is consistently  shorter than the summary, with wh-queries slightly shorter than yes/no queries.

We introduce the novel token percentage: $NTP(string1, string2)$, defined as the percentage of tokens in string1 that are absent in string2. This statistic quantifies the amount of unique information contained in string1 with respect to string2
First, $NTP(doc, query)$ is always lower than $NTP(doc, sum)$, indicating that the generated query always contains less information about the document than the summary.
Subsequently, we observe that $NTP(query, doc)$ is in general higher than $NTP(sum, doc)$, because queries are shorter and contain more unique question words like "what" and "did".
Finally, $NTP(query, sum)$ being considerably  lower than $NTP(sum, query)$ shows that the summary contains more unique information than the query. Furthermore, the query includes a subset of information present in the summary. For instance, a query might inquire about a specific entity in the document, while the summary addresses the query with detailed contexts and facts extracted from the document.

In conclusion, \DatasetNameSpace encompasses documents in both written and spoken languages, covering a wide range of document/summary lengths, abstraction levels, and compression ratios.
\section{\DatasetNameSpace for QFS}
% \usepackage{booktabs}
% \usepackage{graphicx}
\begin{table*}[ht]
\centering
\resizebox{\textwidth}{!}{%
\begin{tabular}{l|l|l}
\hline
Dataset        & Size(Train/Test) & Query Example                                                                                                                   \\ \hline
MultiOpEd      & 1954/560         & is protecting the environment incompatible with capitalism's values?                                                            \\ \hline
NEWTS - word   & 2400/600         & \begin{tabular}[c]{@{}l@{}}snow, weather, cold, winter, temperatures, conditions, hot, morning, \\ expected, parts\end{tabular} \\ \hline
NEWTS - phrase & 2400/600         & \begin{tabular}[c]{@{}l@{}}winter temperatures, hot weather conditions, a cold morning, snow is \\ expected later\end{tabular}  \\ \hline
NEWTS - sentence &
  2400/600 &
  \begin{tabular}[c]{@{}l@{}}This topic is about winter temperatures as opposed to hot weather \\ conditions, cold mornings, and weather forecasts like snow being expected later.\end{tabular} \\ \hline
Debatepedia    & 12000/1000       & \begin{tabular}[c]{@{}l@{}}Would the election of a president make the eu a more accountable \\ institution?\end{tabular}        \\ \hline
DUC 2006 &
  0/200 &
  \begin{tabular}[c]{@{}l@{}}Identify computer viruses detected worldwide.  Include such details \\ as how they are spread, what operating systems they affect, what damage they \\ inflict, their country of origin, and their creators wherever possible.\end{tabular} \\ \hline
DUC 2007 &
  0/180 &
  \begin{tabular}[c]{@{}l@{}}Describe the state of teaching art and music in public schools around the \\ world. Indicate problems, progress and failures.\end{tabular} \\ \hline
% TD-QFS         & 0/120            & alzheimer memory                                                                                                                \\ \hline
\end{tabular}%
}
\caption{Size and example queries of QFS datasets used in evaluation.}
\label{tab:dataset_size_query}
\end{table*}
In this section, we demonstrate that by finetuning pretrained language models on \DatasetName, one can obtain a QFS model that generalizes effectively to unseen tasks and domains. In particular, we finetuned a BART model \cite{lewis-etal-2020-bart}, and the resulting model, \DatasetNameSpace BART, exhibits promising performance on various QFS datasets when directly applied to the unseen test set. Moreover, when extending the fine-tuning process with several thousand in-domain QFS data points, the resulting supervised model surpasses other strong supervised baselines. 

\subsection{Implementation Details}
\label{subsec:impl}
We fine-tuned BART-Large \cite{lewis-etal-2020-bart} on \DatasetNameSpace, using a maximum input length of 1024 and output length of 256. The input string consists of a document and a query, formatted as \textit{question:\textbackslash n <query> \textbackslash n context:\textbackslash n<document>}, where "\textbackslash n" represents a newline character. We employed 8 NVIDIA Tesla V100 GPUs for training, with a batch size of 4 per GPU and an accumulation step of 8, yielding an effective batch size of 256. The BART model was fine-tuned using a learning rate of $3 \times 10^{-5}$ for $50,000$ steps, and the learning rate was scheduled by a polynomial scheduler with $2000$ warmup steps. We set a weight decay of $0.001$ and a label smoothing factor of $0.1$.
For supervised finetuning, we continued to finetune the \DatasetNameSpace BART model with $2000$ total steps and $200$ warm-up steps.
The implementation from Huggingface \cite{wolf-etal-2020-transformers} was utilized.
% \usepackage{booktabs}
% \usepackage{multirow}
% \usepackage{graphicx}
\begin{table*}[ht]
\centering
\resizebox{\textwidth}{!}{%
\begin{tabular}{c|l|llllll}
\hline
\multicolumn{1}{l|}{}              & Model            & ROUGE1 & ROUGE2 & ROUGEL & BERTSCORE & Relevance. Acc. & Stance Acc. \\ \hline
\multirow{6}{*}{Supervised Models} & BART             & 28.2   & 11.3   & 27.0   & 88.7      & 91.9            & 72.3        \\
                                   & +   Rel          & 28.4   & 11.5   & 27.1   & 88.7      & 93.0            & 72.7        \\
                                   & +   Stance       & 28.2   & 11.5   & 26.9   & 88.8      & 91.3            & 73.4        \\
                                   & +   Rel \& Stance & 29.2   & 11.9   & 27.9   & 88.7      & 94.6            & 74.3        \\
                                   & CNN/DM BART      & 26.9   & 10.7   & 25.2   & 86.4      & \textbf{99.5}   & 68.6        \\
 & \DatasetNameSpace BART & \textbf{31.5} & \textbf{13.8} & \textbf{29.8} & \textbf{89.1} & 96.8          & \textbf{78.8} \\ \hline
\multirow{3}{*}{Zero-shot Models}  & InstructGPT 002  & 23.8   & 9.3    & 21.9   & 85.7      & \textbf{99.5}   & 73.2        \\
                                   & CNN/DM BART      & 17.9   & 5.2    & 16.2   & 85.2      & 94.8            & 62.9        \\
 & \DatasetNameSpace BART & \textbf{25.4} & \textbf{10.4} & \textbf{23.6} & \textbf{87.2} & \textbf{99.5} & \textbf{77.0} \\ \hline
\end{tabular}%
}
\caption{ROUGE scores and accuracies of stance and relevance on MultiOpEd dataset. All baseline results except for InstructGPT are
from \cite{liu-etal-2021-multioped}}
\label{tab:multioped_res}
\end{table*}
\subsection{Datasets}
We conduct  evaluation of the finetuned BART-Large model (\DatasetNameSpace BART) on several existing QFS benchmark datasets which have no overlap with \DatasetName.
\begin{itemize}
    \item MultiOpEd \citep{liu-etal-2021-multioped} presents an open-domain news editorial dataset specifically designed to support automatic perspective discovery in news articles. Given a query that explicitly addresses a controversial topic, a system is expected to generate a single-sentence thesis statement that summarizes the arguments presented. Along with ROUGE scores as evaluation metrics, the paper also proposes trained classifiers to assess the correctness and relevance of the generated summary. More specifically, a stance classifier is utilized to predict whether a summary shares the same stance as the news article. For example, a summary that presents an opposing argument to the article might still achieve a high ROUGE score due to n-gram overlap but would receive a low stance accuracy. Similarly, a relevance classifier is employed to evaluate whether the summarized perspective is pertinent to the query.
    \item NEWTS \citep{bahrainian2022newts} dataset is a corpus for summarizing news topics. It is based on the CNN/Dailymail dataset \citep{see-etal-2017-get} and was annotated through online crowd-sourcing. 
    % It is the first dataset that supports the task of topic-focused summarization.
    Each source article is paired with two reference summaries, each focusing on a different theme of the source document. The dataset has 3,000 source articles (2,400 for training, and 600 for testing). In addition to standard ROUGE score, the dataset is evaluated using a LDA topic model indicating the strength of the target topic for the generated summary. We follow the implementation from \citet{bahrainian2022newts} to compute the topic focus score. It is expected that summaries closer to the target topic get higher topic focus scores.
    \item Debatepeida \citep{nema-etal-2017-diversity} was built on Debatepedia - an encyclopedia of pro and con arguments and quotes on critical debate topics. The summaries are highly abstractive and not extractive in the sense that the summary does not necessarily comprise of a sentence which is simply copied or shortened from the original document.
    \item Document Understanding Conferences (DUC) 2006/2007 \footnote{URL at \url{https://www-nlpir.nist.gov/projects/duc}} set up the task to simulate real-world complex question answering. The query in this dataset cannot be answered by simply stating a name, date, quantity, etc. Given a topic and a set of 25 relevant documents, the task is to synthesize a fluent, well-organized 250-word summary of the documents that answers the question(s) in the topic statement.
    % \item TD-QFS \citet{baumel2016topic} prepared a dataset of document clusters in the field of Consumer Health, and asked expert users (Medical Students) to generate summaries of document sets given various queries.
    
\end{itemize}

\subsection{Baselines}
We compare \DatasetNameSpace BART with the following baseline models:
\begin{itemize}
    \item CNN/DM BART is the BART large model finetuned on the query-agnostic CNN/DailyMail \cite{see-etal-2017-get} dataset. This baseline model establishes the lower bound of performance when summarization is solely based on the input document, disregarding the query in the QFS setting.
    \item InstructGPT 002 is the InstructGPT model that can be accessed by directly calling the OpenAI API of \textit{text-davinci-002}. We employ a simple template, "Summarize by answering the following questions:", to link the document with the query and generate content by setting the temperature to 1.0, $top_p=0.9$, and maximum length to 512.
    \item LaQSUM \citep{xu-lapata-2022-document} is a recent model that learns latent queries from documents for abstractive summarization. In contrast to our approach, which explicitly generates the hidden query in natural language, LaQSUM models the query as hidden binary variables to indicate whether a token in the document contributes to the information sought in the summary. The model also requires no QFS annotation and is trained on CNN/DM dataset.
    \item MARGESUM \cite{xu-lapata-2021-generating} is a state-of-the-art few-shot method for QFS which requires a small QFS development set.
    \item GSUM+Query is adapted from GSUM\citep{dou-etal-2021-gsum}, which is a guided summarization system. An unsupervised query-focused extractive system is employed to pre-extract the top-ranked sentences for each test document as guidance. The GSUM model is trained with CNN/DM dataset as well.
    \item QuerySum \cite{xu-lapata-2020-coarse} is an extractive method that utilizes QA datasets as distant supervision to train an evidence estimator for identifying segments likely to answer the query and should be included in the summary.
    \item ProphetNet \cite{qi-etal-2020-prophetnet} is a supervised abstractive summarization model featuring an enhanced objective that predicts the next n tokens simultaneously. We obtain the results of ProphetNet as reported in the NEWTS paper \citep{bahrainian2022newts}.
    \item Unsupervised extractive baselines are taken from \citet{xu-lapata-2022-document}. Lead and LexRank estimate sentence-level centrality using Markov Random Walk on graphs.
    
\end{itemize}

\subsection{Query Unification}

Different QFS datasets have different query formats. For instance, Debatepedia has the query format of a natural question, which is the same as \DatasetName, while the majority of queries in DUC datasets are instructions such as "\textit{Discuss conditions on American Indian reservations or among Native American communities.}" and "\textit{Include the benefits and drawbacks of the reservation system.}". And for NEWTS, the query is a "topic" in the topic model and described in words, phrases or a sentence.

% For TD-QFS, the query is medical noun phrase, such as "Asthma causes" or "Alzheimer's symptoms". For NEWTS, it has three different granularities on query - words, phrase and sentences. The typical query formats for each dataset are shown in table \ref{tab:dataset_size_query}.

To use \DatasetNameSpace in the zero-shot setting, it is necessary to convert the queries of diverse formats into natural questions. Without an off-the-shelf tool for this task, we propose to further utilize \DatasetNameSpace for the query unification task. Specifically, we finetune a BART model to generate queries with the document and summary as input.
Basically, this is what Instruct-GPT was prompted to do in Section \ref{subsec:query_gen}.
%We use the query in \DatasetNameSpace as the target output and a concatenation of the summary and document in the format of \textit{summary:\textbackslash n <summary\textbackslash n context: \textbackslash n <document> >} as the source input.

We denote this finetuned model as $G_{d,s \shortrightarrow q}$ and the finetuned model described in Section \ref{subsec:impl} as  $G_{d,q \shortrightarrow s}$. Given original query $q$ and document $d$, we first use $q$ as a pseudo "summary" and ask $G_{d,s \shortrightarrow q}$ to produce a query $q^{\prime}$ of the desired format, i.e., $q^{\prime} = G_{d,s \shortrightarrow q} (d, q)$. We then use the generated query $q^{\prime}$ as the input query in the follow-up zero-shot inference to predict summary $s = G_{d,q \shortrightarrow s} (d, q^{\prime})$.

The query unification is used to generate quires for NEWTS, DUC 2006, and DUC 2007 dataset. We quantitatively and qualitatively verified its effectiveness in section \ref{subsec:abl_query_unif}.

% We convert such instructions to queries simply by replacing the verb in the instruction to "\textit{What are}" and replacing the trailing period mark to question mark. 
% For TD-QFS, the query is medical noun phrase, such as "Asthma causes" or "Alzheimer's symptoms". We convert them to query sentences similarly by adding "\textit{What are}" before the phrase and a question mark after. 
% \usepackage{booktabs}
% \usepackage{multirow}
% \usepackage{graphicx}
\begin{table*}[]
\centering
\resizebox{\textwidth}{!}{%
\begin{tabular}{c|lllll}
\hline
\multicolumn{1}{l|}{Category}                  & Model                                     & R-1           & R-2           & R-L           & Topic Score    \\ \hline
\multirow{9}{*}{Supervised}                    & ProphetNet supervised w/ topic words      & 31.9          & 10.8          & 20.7          & 0.136          \\
                                               & ProphetNet supervised w/ topic phrases    & 31.6          & 10.4          & 20.2          & 0.147          \\
                                               & ProphetNet supervised w/ topic sentences  & 31.4          & 10.0          & 20.0          & 0.163          \\
                                               & CNN/DM BART w/ topic words                & 34.5          & 11.5          & 22.1          & 0.176          \\
                                               & CNN/DM BART w/ topic phrases              & 34.3          & 11.3          & 22.0          & 0.178          \\
                                               & CNN/DM BART w/ topic sentences            & 34.5          & 11.5          & 22.0          & 0.174          \\
                                               & \DatasetNameSpace BART w/ topic words     & 34.4          & 11.9          & 22.5          & 0.178          \\
 & \DatasetNameSpace BART w/ topic phrases                                                       & \textbf{34.6} & \textbf{12.0} & \textbf{22.8} & \textbf{0.180} \\
 & \DatasetNameSpace BART w/ topic sentences                                                     & \textbf{34.6} & \textbf{12.0} & 22.7          & \textbf{0.180} \\ \hline
\multirow{7}{*}{Zero-shot/Transfer   Learning} & CNN/DM BART                               & 31.2          & 10.4          & 20.8          & 0.125          \\
                                               & Plug and Play Language Model              & 29.6          & 9.1           & 18.8          & 0.148          \\
 & \begin{tabular}[c]{@{}l@{}}Customizable Abstractive Topic-based \\ Summarization\end{tabular} & 30.1          & 9.4           & 19.1          & 0.152          \\
                                               & InstructGPT 002                           & 32.7          & 10.7          & 21.3          & 0.184          \\
                                               & \DatasetNameSpace BART w/ topic words     & \textbf{33.3} & \textbf{11.2} & \textbf{21.6} & 0.141          \\
                                               & \DatasetNameSpace BART w/ topic phrases   & 32.6          & 10.8          & 21.0          & 0.161          \\
                                               & \DatasetNameSpace BART w/ topic sentences & 32.4          & 10.5          & 20.9          & \textbf{0.187} \\ \hline
\multicolumn{1}{l|}{}                          & Ground Truth                              & —             & —             & —             & 0.193          \\ \hline
\end{tabular}%
}
\caption{ROUGE scores and topic scores on NEWTS dataset. All baseline results except for InstructGPT 002 are from \citep{bahrainian2022newts}.}
\label{tab:newts_res}
\end{table*}
\subsection{Mutli-document Query Focused Summarization}

Since \DatasetNameSpace contains only single-document QFS data, the fine-tuned model $G_{d,q \shortrightarrow s}$ can generate summaries based on individual document-query pairs. To evaluate zero-shot multi-document QFS, we adopt a straightforward iterative approach from previous works by \citet{baumel2018query,xu-lapata-2022-document}. Given a cluster of documents and a query, we first rank documents using term frequency-inverse document frequency, then generate a summary for each ranked document. The final summary is chosen from the top-ranked list. Following the list order, we successively concatenate a summary if its token overlap percentage with the selected summaries is below a threshold, e.g., $50\%$, until the total length of chosen summaries reaches a predefined token budget (e.g., 250 tokens).
% original one is below. the above one is from GPT-4 editing
%As \DatasetNameSpace only contains single-document QFS data, the finetuned model $G_{d,q \shortrightarrow s}$ can only generate a summary based on a pair of single document and query. To facilitate the evaluation on zero-shot multi-document QFS, we followed previous works of \citet{baumel2018query,xu-lapata-2022-document} for a simple iterative approach. Given the cluster of documents and a given query, we first rank the documents via term frequency-inverse document frequency, and then generate a summary for each document in the ranked list. The final summary is selected from the top of the ranked list. In the order of the list, we sequentially concatenate a summary if its percentage of overlapped tokens with selected summaries is lower than a threshold, e.g., $50\%$, until the total length of selected summaries reach a pre-defined token budget (e.g. 250 tokens). 

\section{Evaluation Result}
% \input{tables/multioped_res}

% \usepackage{booktabs}
% \usepackage{multirow}
% \usepackage{graphicx}
\begin{table}[ht]
\centering
\resizebox{0.5\textwidth}{!}{%
\begin{tabular}{@{}cllll@{}}
\toprule
\multicolumn{1}{l}{Category} & Model           & R-1  & R-2  & R-L  \\ \midrule
\multicolumn{1}{c|}{\multirow{2}{*}{Unsupervised}}                  & LEAD                   & 18.1          & 5.6          & 15.9          \\
\multicolumn{1}{c|}{}        & LexRank         & 17.4 & 5.3  & 15.1 \\ \midrule
\multicolumn{1}{c|}{\multirow{3}{*}{Supervised}}                    & DDA                    & 7.4           & 2.8          & 7.2           \\
\multicolumn{1}{c|}{}        & BERTAbs+Rank    & 19.2 & 10.6 & 17.9 \\
\multicolumn{1}{c|}{}        & BERTAbs+Concat  & 26.4 & 11.9 & 25.1 \\ \midrule
\multicolumn{1}{c|}{\multirow{6}{*}{Zero-shot/Transfer   Learning}} & BERTAbs                & 13.3          & 2.8          & 2.8           \\
\multicolumn{1}{c|}{}        & CNN/DM BART     & 21.4 & 6.3  & 18.4 \\
\multicolumn{1}{c|}{}        & InstructGPT 002 & 21.8 & 6.5  & 18.8 \\
\multicolumn{1}{c|}{}        & GSUM+Query      & 21.2 & 6.2  & 18.2 \\
\multicolumn{1}{c|}{}        & LaQSUM          & 23.5 & 7.2  & 20.6 \\
\multicolumn{1}{c|}{}                                               & \DatasetNameSpace BART & \textbf{23.6} & \textbf{7.6} & \textbf{21.0} \\ \bottomrule
\end{tabular}%
}
\caption{ROUGE scores on single-document QFS dataset Debatepedia. Baseline results (except for InstructGPT 002) are reported from \citet{xu-lapata-2022-document}.}
\label{tab:debatepedia_res}
\end{table}
% \cz{need a subsection talking about baseline models}
\subsection{Results on Single-document QFS}

The table \ref{tab:multioped_res} presents the ROUGE scores and accuracies of stance and relevance for various models on the MultiOpEd dataset. It can be observed that \DatasetNameSpace BART outperforms other models in both supervised and unsupervised cases, achieving the highest ROUGE scores and stance accuracies in both settings. For relevance accuracy, it also achieves the best in the zero-shot setting and the second best in the supervised setting. This demonstrates the robust performance of \DatasetNameSpace BART across different settings. Interestingly, in the supervised setting, pre-finetuning on the CNN/DailyMail dataset (CNN/DM BART) actually diminishes performance compared to vanilla BART without pre-finetuning. This result indicates that a general summarization dataset may not always be beneficial for QFS and highlights the necessity for high-quality, large-scale QFS datasets like \DatasetNameSpace.

Similarly, Table \ref{tab:newts_res} presents the ROUGE scores (R-1, R-2, and R-L) and topic scores on the NEWTS dataset for different models under two categories: Supervised and Zero-shot/Transfer Learning.
We used "w/ \{query\_granularity\}" to denote the results using three different granularities for the query: words, phrases, and sentences. For instance, "ProphetNet supervised w/ topic words" refers to the result ProphetNet achieved using a query of topic words.
Overall, the \DatasetNameSpace BART models outperform other baselines in terms of ROUGE scores, with the \DatasetNameSpace BART w/ topic words model achieving the highest scores in the zero-shot setting and the \DatasetNameSpace BART w/ topic phrases model obtaining the best results in the supervised setting. Additionally, the \DatasetNameSpace BART w/ topic sentences model achieves the highest topic score among all models in both zero-shot and supervised settings, closely approaching the topic scores of the ground truth. Without fine-tuning on any supervised data, \DatasetNameSpace BART exhibits a significant advantage over the supervised ProphetNet models in terms of ROUGE scores and topic scores.
The supervised results also reveal that \DatasetNameSpace remains more beneficial even when some in-domain supervised data (2,400 training samples from NEWTS) is accessible.

% \cz{need to explain what is \DatasetNameSpace BART w/ different topic components}

% Table \ref{tab:debatepedia_res} shows the ROUGE scores on the single-document QFS dataset Debatepedia for various models categorized as unsupervised, supervised, and zero-shot/transfer learning. The standout performance in this table is of the \DatasetNameSpace BART model, which belongs to the zero-shot/transfer learning category. It achieved the highest ROUGE scores of 23.6 for R-1, 7.6 for R-2, and 21.0 for R-L, outperforming all other models in the category of zero-shot/transfer learning.

% GPT-4 edited
Table \ref{tab:debatepedia_res} presents the ROUGE scores on the single-document QFS dataset Debatepedia for various models, classified into unsupervised, supervised, and zero-shot/transfer learning categories. \DatasetNameSpace BART achieves the highest ROUGE scores,
% of 23.6 for R-1, 7.6 for R-2, and 21.0 for R-L,
surpassing all other models in the zero-shot/transfer learning category.

It is worth mentioning that our model distilled from InstructGPT outperforms the teacher model in the all single-document QFS datasets.

% A recent work of \citet{10.1162/coli_a_00434} found some queries do not have any relations with the input documents. To this end, we conduct a human study to compare the system output of \DatasetNameSpace BART model with the reference from Debatepedia. The human annotators are instructed to select the better summary out of two candidate given the query and the context document. If two summaries are of equal quality, or the query is not answerable from the document, we will mark them as "Tie". In the blind test, human annotators prefer the output of \DatasetNameSpace BART model 16 times, the reference 9 times and select "Tie" for 16 times. This shows that \DatasetNameSpace is essentially of higher quality than existing benchmark dataset like Debatepedia.

% GPT-4 edited
\subsubsection{Human Study}
A recent study by \citet{10.1162/coli_a_00434} discovered that some queries have no relation to the input documents. To investigate this, we conducted a human study comparing the \DatasetNameSpace BART model's output with the Debatepedia reference. Human annotators were instructed to choose the better summary from two candidates, given the query and the context document. If both summaries were of equal quality or the query was unanswerable from the document, they would mark it as a "Tie." In the blind test, annotators preferred the \DatasetNameSpace BART model's output 16 times, the reference 9 times, and selected "Tie" 16 times. This indicates that \DatasetNameSpace has a higher quality compared to existing benchmark datasets like Debatepedia. Additionally, we observed that model finetuned on our dataset will summarize the document and try to answer the question by giving a direct answer of supporting or opposing the statement in the query.

% \cz{need ablation study to show the effect of $G_{d,q \shortrightarrow s}$}

% \usepackage{booktabs}
% \usepackage{multirow}
% \usepackage{graphicx}
\begin{table*}[ht]
\centering
\resizebox{0.8\textwidth}{!}{%
\begin{tabular}{@{}c|l|lll|lll@{}}
\toprule
\multicolumn{1}{l|}{}                 &                                       & \multicolumn{3}{c|}{DUC 2006} & \multicolumn{3}{c}{DUC 2007} \\ \midrule
\multicolumn{1}{l|}{Category}         & Model                                 & R-1      & R-2      & R-L     & R-1      & R-2     & R-L     \\ \midrule
\multirow{4}{*}{Upper Bound \&   Baselines}      & Gold            & 45.4          & 11.2         & 16.8          & 47.5          & 14.0          & 18.9          \\
                                      & Oracle                                & 47.5     & 15.8     & 20.2    & 47.6     & 17.1    & 20.9    \\
                                      & Lead                                  & 32.1     & 5.3      & 10.4    & 33.4     & 6.5     & 11.3    \\
                                      & LexRank                               & 34.2     & 6.4      & 11.4    & 35.8     & 7.7     & 12.7    \\ \midrule
\multirow{3}{*}{Distantly Supervised} & QuerySum                              & 41.6     & 9.5      & 15.3    & 43.3     & 11.6    & 16.8    \\
                                      & Bart-CAQ                              & 38.3     & 7.7      & 12.9    & 40.5     & 9.2     & 14.4    \\
                                      & PQSum                                 & 40.9     & 9.4      & 14.8    & 42.2     & 10.8    & 16.0    \\ \midrule
\multirow{6}{*}{Few- or Zero-shot   Abstractive} & MARGESUM*       & 40.2          & 9.7          & 15.1          & 42.5          & 12.0          & 16.9          \\
                                      & CNN/DM BART                           & 38.3     & 7.8      & 13.1    & 40.2     & 9.9     & 14.6    \\
                                      & GSUM+Query                            & 38.1     & 7.9      & 13.1    & 39.5     & 9.5     & 14.3    \\
                                      & LQSUM                                 & 39.1     & 8.5      & 13.7    & 40.4     & 10.2    & 15.0    \\
                                                 & InstructGPT 002 & \textbf{41.1} & \textbf{9.4} & \textbf{15.3} & \textbf{42.0} & \textbf{10.7} & \textbf{16.1} \\
                                      & \DatasetNameSpace BART & 40.0     & 9.0      & 14.0    & 41.3     & 10.6    & 15.3    \\ \bottomrule
\end{tabular}%
}
\caption{ROUGE scores on multi-document QFS dataset DUC 2006, DUC2007. "*" means the model is few-shot instead of zero-shot. Baseline results (except for InstructGPT 002) are reported from \citet{xu-lapata-2022-document}.}
\label{tab:mds_res}
\end{table*}
\subsection{Results on Multi-document QFS}
Table \ref{tab:mds_res} presents ROUGE scores for various summarization models on multi-document QFS datasets, namely DUC 2006, DUC 2007. The models are categorized into Upper Bound \& Baselines, Distantly Supervised, and Few- or Zero-shot Abstractive categories. Note that MARGESUM in this category is a few-shot model while the others are zero-shot ones.

Among the zero-shot abstractive models, \DatasetNameSpace BART exhibits the second-best performance in terms of ROUGE scores on both DUC 2006 and DUC 2007 benchmarks, only trailing behind its teacher model, InstructGPT 002. We hypothesize that the primary reason for this is the prevalence of queries in the DUC datasets that are presented in a human-readable instruction format, which inherently favors the instruction-following nature of InstructGPT. Despite being a considerably smaller model, \DatasetNameSpace BART still demonstrates promising instruction-following capabilities by leveraging our query unification method.

% While it does not achieve the highest scores on TD-QFS, it still exhibits competitive performance, showcasing its robust generalization capability across various QFS tasks. 
% \cz{a bit wordy in the results here, as people can see the results from table. Can highlight how much higher ROUGE LMGQS BART obtains}
% Specifically, it scores 41.1 in ROUGE-1 and ROUGE-2, and 14.8 in ROUGE-L for DUC 2006, while for DUC 2007, it scores 41.1, 10.7, and 15.6 in ROUGE-1, ROUGE-2, and ROUGE-L, respectively. However, on the TD-QFS dataset, it doesn't outperform LQSUM, which has the highest scores of 45.7, 18.1, and 22.1 in ROUGE-1, ROUGE-2, and ROUGE-L, respectively. Overall, \DatasetNameSpace BART demonstrates strong performance in summarization tasks across these datasets.

\begin{figure}[ht]
    \centering
    \resizebox{0.5\textwidth}{!}{%
    \includegraphics{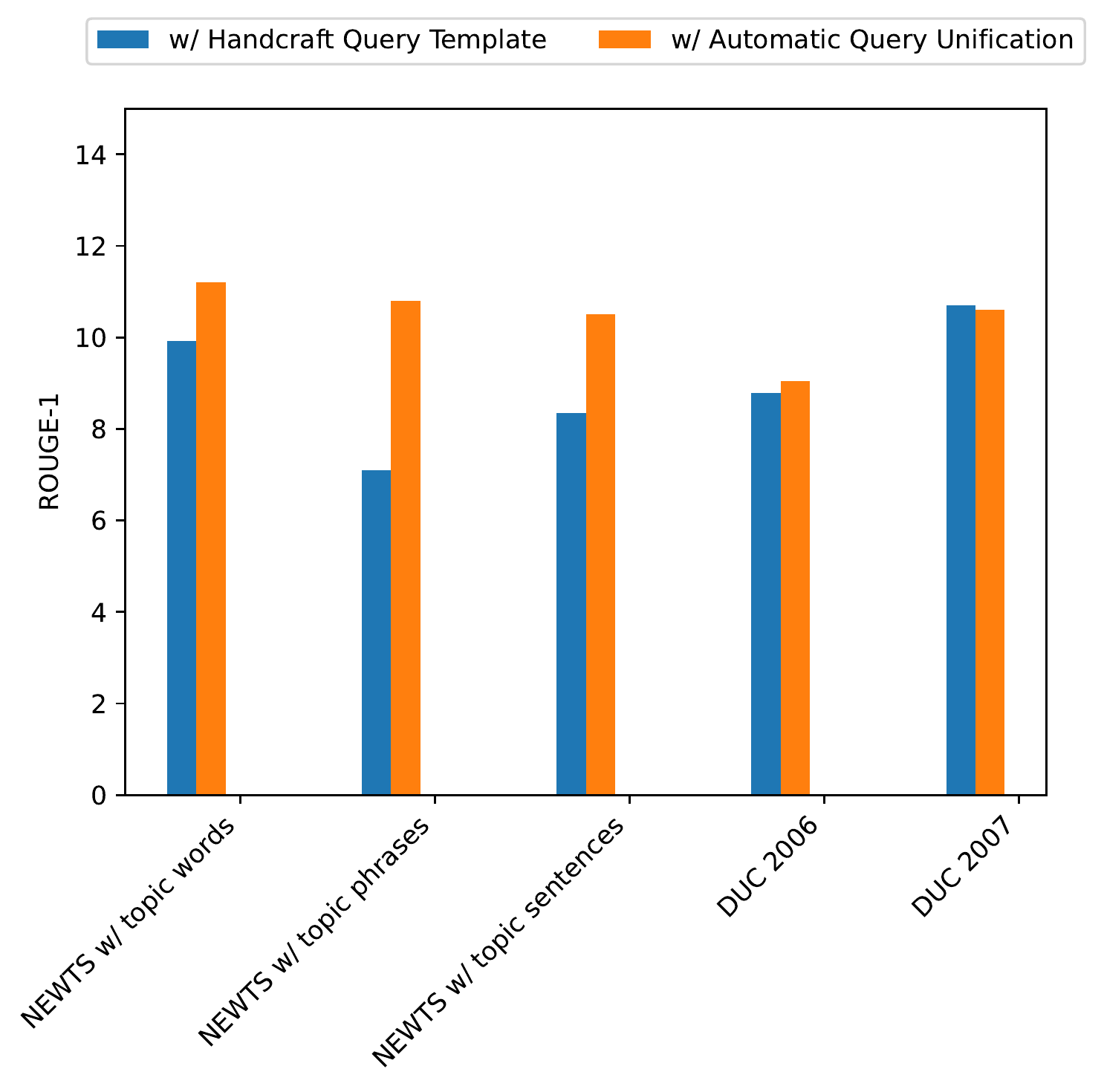}
    }
    \caption{Ablation study on the effect of query unification. For simplicity, we only present ROUGE-2 score in this figure.}
    \label{fig:query_uni_abl}
\end{figure}

\subsection{Ablation Study of Query Unification}
\label{subsec:abl_query_unif}
To evaluate the efficacy of query unification, we perform an ablation study to compare the quality of automatically generated queries $q^{\prime} = G_{d,s \shortrightarrow q} (d, q)$. For comparison, we manually create a query template to transform the query into a natural language question. The template is selected separately for the NEWTS and DUC datasets, and the authors utilize the generation on the development set of these datasets to carefully refine the template. In Figure \ref{fig:query_uni_abl}, we present a comparison of ROUGE-2 scores between \DatasetNameSpace BART when employing 1) manually crafted query templates or 2) automatically generated queries from query unification. It is evident that query unification holds an advantage over the handcrafted template, despite the latter necessitating access to a validation set and meticulous tuning from human experts.

\section{Summary}
We introduce a novel dataset, \DatasetNameSpace, for query-focused summarization (QFS), addressing the scarcity of large-scale benchmarks in this domain. In order to reduce the laborious human effort required for QFS, we utilize a large-scale language model to generate hidden queries from existing query-agnostic summarization datasets. By performing standard finetuning on \DatasetNameSpace, we attain state-of-the-art zero-shot performance across multiple QFS benchmarks. The resulting zero-shot model can be further enhanced by finetuning on labeled QFS datasets, achieving the state-of-the-art in supervised performance. In addition to QFS, another promising avenue for future research is open-domain question answering (QA). Considering the typically extensive context, a query-focused summary of the context could prove advantageous for downstream QA models.

\bibliography{anthology,custom}

\begin{thebibliography}{27}
\expandafter\ifx\csname natexlab\endcsname\relax\def\natexlab#1{#1}\fi

\bibitem[{Bahrainian et~al.(2022)Bahrainian, Feucht, and
  Eickhoff}]{bahrainian2022newts}
Seyed~Ali Bahrainian, Sheridan Feucht, and Carsten Eickhoff. 2022.
\newblock Newts: A corpus for news topic-focused summarization.
\newblock In \emph{Findings of the Association for Computational Linguistics:
  ACL 2022}, pages 493--503.

\bibitem[{Baumel et~al.(2016)Baumel, Cohen, and Elhadad}]{baumel2016topic}
Tal Baumel, Raphael Cohen, and Michael Elhadad. 2016.
\newblock Topic concentration in query focused summarization datasets.
\newblock In \emph{Thirtieth AAAI Conference on Artificial Intelligence}.

\bibitem[{Baumel et~al.(2018)Baumel, Eyal, and Elhadad}]{baumel2018query}
Tal Baumel, Matan Eyal, and Michael Elhadad. 2018.
\newblock Query focused abstractive summarization: Incorporating query
  relevance, multi-document coverage, and summary length constraints into
  seq2seq models.
\newblock \emph{arXiv preprint arXiv:1801.07704}.

\bibitem[{Chen et~al.(2021)Chen, Liu, Chen, and
  Zhang}]{chen-etal-2021-dialogsum}
Yulong Chen, Yang Liu, Liang Chen, and Yue Zhang. 2021.
\newblock \href {https://doi.org/10.18653/v1/2021.findings-acl.449}
  {{D}ialog{S}um: {A} real-life scenario dialogue summarization dataset}.
\newblock In \emph{Findings of the Association for Computational Linguistics:
  ACL-IJCNLP 2021}, pages 5062--5074, Online. Association for Computational
  Linguistics.

\bibitem[{Dang(2006{\natexlab{a}})}]{dang-2006-duc}
Hoa~Trang Dang. 2006{\natexlab{a}}.
\newblock \href {https://aclanthology.org/W06-0707} {{DUC} 2005: Evaluation of
  question-focused summarization systems}.
\newblock In \emph{Proceedings of the Workshop on Task-Focused Summarization
  and Question Answering}, pages 48--55, Sydney, Australia. Association for
  Computational Linguistics.

\bibitem[{Dang(2006{\natexlab{b}})}]{dang2006duc}
Hoa~Trang Dang. 2006{\natexlab{b}}.
\newblock Duc 2005: Evaluation of question-focused summarization systems.
\newblock In \emph{Proceedings of the Workshop on Task-Focused Summarization
  and Question Answering}, pages 48--55.

\bibitem[{Dou et~al.(2021)Dou, Liu, Hayashi, Jiang, and
  Neubig}]{dou-etal-2021-gsum}
Zi-Yi Dou, Pengfei Liu, Hiroaki Hayashi, Zhengbao Jiang, and Graham Neubig.
  2021.
\newblock \href {https://doi.org/10.18653/v1/2021.naacl-main.384} {{GS}um: A
  general framework for guided neural abstractive summarization}.
\newblock In \emph{Proceedings of the 2021 Conference of the North American
  Chapter of the Association for Computational Linguistics: Human Language
  Technologies}, pages 4830--4842, Online. Association for Computational
  Linguistics.

\bibitem[{Gehrmann et~al.(2018)Gehrmann, Deng, and
  Rush}]{gehrmann-etal-2018-bottom}
Sebastian Gehrmann, Yuntian Deng, and Alexander Rush. 2018.
\newblock \href {https://doi.org/10.18653/v1/D18-1443} {Bottom-up abstractive
  summarization}.
\newblock In \emph{Proceedings of the 2018 Conference on Empirical Methods in
  Natural Language Processing}, pages 4098--4109, Brussels, Belgium.
  Association for Computational Linguistics.

\bibitem[{Gliwa et~al.(2019)Gliwa, Mochol, Biesek, and
  Wawer}]{gliwa-etal-2019-samsum}
Bogdan Gliwa, Iwona Mochol, Maciej Biesek, and Aleksander Wawer. 2019.
\newblock \href {https://doi.org/10.18653/v1/D19-5409} {{SAMS}um corpus: A
  human-annotated dialogue dataset for abstractive summarization}.
\newblock In \emph{Proceedings of the 2nd Workshop on New Frontiers in
  Summarization}, pages 70--79, Hong Kong, China. Association for Computational
  Linguistics.

\bibitem[{Laskar et~al.(2022)Laskar, Hoque, and Huang}]{10.1162/coli_a_00434}
Md~Tahmid~Rahman Laskar, Enamul Hoque, and Jimmy~Xiangji Huang. 2022.
\newblock \href {https://doi.org/10.1162/coli_a_00434} {{Domain Adaptation with
  Pre-trained Transformers for Query-Focused Abstractive Text Summarization}}.
\newblock \emph{Computational Linguistics}, 48(2):279--320.

\bibitem[{Lewis et~al.(2020)Lewis, Liu, Goyal, Ghazvininejad, Mohamed, Levy,
  Stoyanov, and Zettlemoyer}]{lewis-etal-2020-bart}
Mike Lewis, Yinhan Liu, Naman Goyal, Marjan Ghazvininejad, Abdelrahman Mohamed,
  Omer Levy, Veselin Stoyanov, and Luke Zettlemoyer. 2020.
\newblock \href {https://doi.org/10.18653/v1/2020.acl-main.703} {{BART}:
  Denoising sequence-to-sequence pre-training for natural language generation,
  translation, and comprehension}.
\newblock In \emph{Proceedings of the 58th Annual Meeting of the Association
  for Computational Linguistics}, pages 7871--7880, Online. Association for
  Computational Linguistics.

\bibitem[{Liu et~al.(2021)Liu, Chen, Uyttendaele, and
  Roth}]{liu-etal-2021-multioped}
Siyi Liu, Sihao Chen, Xander Uyttendaele, and Dan Roth. 2021.
\newblock \href {https://doi.org/10.18653/v1/2021.naacl-main.344}
  {{M}ulti{O}p{E}d: A corpus of multi-perspective news editorials}.
\newblock In \emph{Proceedings of the 2021 Conference of the North American
  Chapter of the Association for Computational Linguistics: Human Language
  Technologies}, pages 4345--4361, Online. Association for Computational
  Linguistics.

\bibitem[{Liu and Lapata(2019)}]{liu-lapata-2019-text}
Yang Liu and Mirella Lapata. 2019.
\newblock \href {https://doi.org/10.18653/v1/D19-1387} {Text summarization with
  pretrained encoders}.
\newblock In \emph{Proceedings of the 2019 Conference on Empirical Methods in
  Natural Language Processing and the 9th International Joint Conference on
  Natural Language Processing (EMNLP-IJCNLP)}, pages 3730--3740, Hong Kong,
  China. Association for Computational Linguistics.

\bibitem[{Nallapati et~al.(2016)Nallapati, Zhou, dos Santos, Gu̇l{\c{c}}ehre,
  and Xiang}]{nallapati-etal-2016-abstractive}
Ramesh Nallapati, Bowen Zhou, Cicero dos Santos, {\c{C}}a{\u{g}}lar
  Gu̇l{\c{c}}ehre, and Bing Xiang. 2016.
\newblock \href {https://doi.org/10.18653/v1/K16-1028} {Abstractive text
  summarization using sequence-to-sequence {RNN}s and beyond}.
\newblock In \emph{Proceedings of The 20th {SIGNLL} Conference on Computational
  Natural Language Learning}, pages 280--290, Berlin, Germany. Association for
  Computational Linguistics.

\bibitem[{Narayan et~al.(2018)Narayan, Cohen, and
  Lapata}]{narayan-etal-2018-dont}
Shashi Narayan, Shay~B. Cohen, and Mirella Lapata. 2018.
\newblock \href {https://doi.org/10.18653/v1/D18-1206} {Don{'}t give me the
  details, just the summary! topic-aware convolutional neural networks for
  extreme summarization}.
\newblock In \emph{Proceedings of the 2018 Conference on Empirical Methods in
  Natural Language Processing}, pages 1797--1807, Brussels, Belgium.
  Association for Computational Linguistics.

\bibitem[{Nema et~al.(2017)Nema, Khapra, Laha, and
  Ravindran}]{nema-etal-2017-diversity}
Preksha Nema, Mitesh~M. Khapra, Anirban Laha, and Balaraman Ravindran. 2017.
\newblock \href {https://doi.org/10.18653/v1/P17-1098} {Diversity driven
  attention model for query-based abstractive summarization}.
\newblock In \emph{Proceedings of the 55th Annual Meeting of the Association
  for Computational Linguistics (Volume 1: Long Papers)}, pages 1063--1072,
  Vancouver, Canada. Association for Computational Linguistics.

\bibitem[{Ouyang et~al.(2022)Ouyang, Wu, Jiang, Almeida, Wainwright, Mishkin,
  Zhang, Agarwal, Slama, Ray et~al.}]{ouyang2022training}
Long Ouyang, Jeff Wu, Xu~Jiang, Diogo Almeida, Carroll~L Wainwright, Pamela
  Mishkin, Chong Zhang, Sandhini Agarwal, Katarina Slama, Alex Ray, et~al.
  2022.
\newblock Training language models to follow instructions with human feedback.
\newblock \emph{arXiv preprint arXiv:2203.02155}.

\bibitem[{Qi et~al.(2020)Qi, Yan, Gong, Liu, Duan, Chen, Zhang, and
  Zhou}]{qi-etal-2020-prophetnet}
Weizhen Qi, Yu~Yan, Yeyun Gong, Dayiheng Liu, Nan Duan, Jiusheng Chen, Ruofei
  Zhang, and Ming Zhou. 2020.
\newblock \href {https://doi.org/10.18653/v1/2020.findings-emnlp.217}
  {{P}rophet{N}et: Predicting future n-gram for
  sequence-to-{S}equence{P}re-training}.
\newblock In \emph{Findings of the Association for Computational Linguistics:
  EMNLP 2020}, pages 2401--2410, Online. Association for Computational
  Linguistics.

\bibitem[{See et~al.(2017)See, Liu, and Manning}]{see-etal-2017-get}
Abigail See, Peter~J. Liu, and Christopher~D. Manning. 2017.
\newblock \href {https://doi.org/10.18653/v1/P17-1099} {Get to the point:
  Summarization with pointer-generator networks}.
\newblock In \emph{Proceedings of the 55th Annual Meeting of the Association
  for Computational Linguistics (Volume 1: Long Papers)}, pages 1073--1083,
  Vancouver, Canada. Association for Computational Linguistics.

\bibitem[{Sutskever et~al.(2014)Sutskever, Vinyals, and Le}]{NIPS2014_a14ac55a}
Ilya Sutskever, Oriol Vinyals, and Quoc~V Le. 2014.
\newblock \href
  {https://proceedings.neurips.cc/paper/2014/file/a14ac55a4f27472c5d894ec1c3c743d2-Paper.pdf}
  {Sequence to sequence learning with neural networks}.
\newblock In \emph{Advances in Neural Information Processing Systems},
  volume~27. Curran Associates, Inc.

\bibitem[{Vaswani et~al.(2017)Vaswani, Shazeer, Parmar, Uszkoreit, Jones,
  Gomez, Kaiser, and Polosukhin}]{NIPS2017_3f5ee243}
Ashish Vaswani, Noam Shazeer, Niki Parmar, Jakob Uszkoreit, Llion Jones,
  Aidan~N Gomez, \L~ukasz Kaiser, and Illia Polosukhin. 2017.
\newblock \href
  {https://proceedings.neurips.cc/paper/2017/file/3f5ee243547dee91fbd053c1c4a845aa-Paper.pdf}
  {Attention is all you need}.
\newblock In \emph{Advances in Neural Information Processing Systems},
  volume~30. Curran Associates, Inc.

\bibitem[{Wolf et~al.(2020)Wolf, Debut, Sanh, Chaumond, Delangue, Moi, Cistac,
  Rault, Louf, Funtowicz, Davison, Shleifer, von Platen, Ma, Jernite, Plu, Xu,
  Le~Scao, Gugger, Drame, Lhoest, and Rush}]{wolf-etal-2020-transformers}
Thomas Wolf, Lysandre Debut, Victor Sanh, Julien Chaumond, Clement Delangue,
  Anthony Moi, Pierric Cistac, Tim Rault, Remi Louf, Morgan Funtowicz, Joe
  Davison, Sam Shleifer, Patrick von Platen, Clara Ma, Yacine Jernite, Julien
  Plu, Canwen Xu, Teven Le~Scao, Sylvain Gugger, Mariama Drame, Quentin Lhoest,
  and Alexander Rush. 2020.
\newblock \href {https://doi.org/10.18653/v1/2020.emnlp-demos.6} {Transformers:
  State-of-the-art natural language processing}.
\newblock In \emph{Proceedings of the 2020 Conference on Empirical Methods in
  Natural Language Processing: System Demonstrations}, pages 38--45, Online.
  Association for Computational Linguistics.

\bibitem[{Xu and Lapata(2020)}]{xu-lapata-2020-coarse}
Yumo Xu and Mirella Lapata. 2020.
\newblock \href {https://doi.org/10.18653/v1/2020.emnlp-main.296}
  {Coarse-to-fine query focused multi-document summarization}.
\newblock In \emph{Proceedings of the 2020 Conference on Empirical Methods in
  Natural Language Processing (EMNLP)}, pages 3632--3645, Online. Association
  for Computational Linguistics.

\bibitem[{Xu and Lapata(2021)}]{xu-lapata-2021-generating}
Yumo Xu and Mirella Lapata. 2021.
\newblock \href {https://doi.org/10.18653/v1/2021.acl-long.475} {Generating
  query focused summaries from query-free resources}.
\newblock In \emph{Proceedings of the 59th Annual Meeting of the Association
  for Computational Linguistics and the 11th International Joint Conference on
  Natural Language Processing (Volume 1: Long Papers)}, pages 6096--6109,
  Online. Association for Computational Linguistics.

\bibitem[{Xu and Lapata(2022)}]{xu-lapata-2022-document}
Yumo Xu and Mirella Lapata. 2022.
\newblock \href {https://doi.org/10.1162/tacl_a_00480} {Document summarization
  with latent queries}.
\newblock \emph{Transactions of the Association for Computational Linguistics},
  10:623--638.

\bibitem[{Zhong et~al.(2021)Zhong, Yin, Yu, Zaidi, Mutuma, Jha, Awadallah,
  Celikyilmaz, Liu, Qiu, and Radev}]{zhong-etal-2021-qmsum}
Ming Zhong, Da~Yin, Tao Yu, Ahmad Zaidi, Mutethia Mutuma, Rahul Jha,
  Ahmed~Hassan Awadallah, Asli Celikyilmaz, Yang Liu, Xipeng Qiu, and Dragomir
  Radev. 2021.
\newblock \href {https://doi.org/10.18653/v1/2021.naacl-main.472} {{QMS}um: A
  new benchmark for query-based multi-domain meeting summarization}.
\newblock In \emph{Proceedings of the 2021 Conference of the North American
  Chapter of the Association for Computational Linguistics: Human Language
  Technologies}, pages 5905--5921, Online. Association for Computational
  Linguistics.

\bibitem[{Zhu et~al.(2021)Zhu, Liu, Mei, and Zeng}]{zhu-etal-2021-mediasum}
Chenguang Zhu, Yang Liu, Jie Mei, and Michael Zeng. 2021.
\newblock \href {https://doi.org/10.18653/v1/2021.naacl-main.474}
  {{M}edia{S}um: A large-scale media interview dataset for dialogue
  summarization}.
\newblock In \emph{Proceedings of the 2021 Conference of the North American
  Chapter of the Association for Computational Linguistics: Human Language
  Technologies}, pages 5927--5934, Online. Association for Computational
  Linguistics.

\end{thebibliography}

\appendix

\section{Appendix}
% \usepackage{booktabs}
% \usepackage{graphicx}
\begin{table*}[]
\centering
\resizebox{\textwidth}{!}{%
\begin{tabular}{@{}lll@{}}
\toprule
Domain &
  News &
  Dialogue \\ \midrule
Document &
  \begin{tabular}[c]{@{}l@{}}BOGOTA, Colombia (CNN) -- A key rebel commander and fugitive \\ from a U.S. drug trafficking indictment was killed over the weekend in an air \\ attack on a guerrilla encampment, the Colombian military said Monday. Alleged  \\ cocaine trafficker and FARC rebel Tomas Medina Caracas in an Interpol photo.  \\ Tomas Medina Caracas, known popularly as \textbackslash{}"El Negro Acacio,\textbackslash{}" was a \\ member of the high command of the Fuerzas Armadas Revolucionarias de Colombia \\ and, according to Colombian and U.S. officials, helped manage the group's \\ extensive cocaine trafficking network. He had been in the cross-hairs of the \\ U.S. Justice Department since 2002. He was charged with conspiracy to import \\ cocaine into the United States and manufacturing and distributing cocaine \\ within Colombia to fund the FARC's 42-year insurgency against the government. \\ U.S. officials alleged Medina Caracas managed the rebel group's sales of  \\ cocaine to international drug traffickers, who in turn smuggled it into the \\ United States. He was also indicted in the United States along with two other \\ FARC commanders in November 2002 on charges of conspiring to kidnap two U.S. \\ oil workers from neighboring Venezuela in 1997 and holding one of them for \\ nine months until a \$1 million ransom was paid. Officials said the army's  \\ Rapid Response Force, backed by elements of the Colombian Air Force, tracked  \\ Medina Caracas down at a FARC camp in the jungle in the south of the country. \\ \textbackslash{}"After a bombardment, the troops occupied the camp, and they've found \\ 14 dead rebels so far, along with rifles, pistols, communications equipment \\ and ... four GPS systems,\textbackslash{}" Defense Minister Juan Manuel Santos said at \\ a news conference. \textbackslash{}"The death of 'El Negro Acacio' was confirmed by \\ various sources, including members of FARC itself.\textbackslash{}" Medina Caracas \\ commanded FARC's 16th Front in the southern departments of Vichada and \\ Guainia. Established in 1964 as the military wing of the Colombian Communist \\ Party, FARC is Colombia's oldest, largest, most capable and best-equipped \\ Marxist rebel group, according to the U.S. Department of State. E-mail to a \\ friend . Journalist Fernando Ramos contributed to this report.\end{tabular} &
  \begin{tabular}[c]{@{}l@{}}Emma: I\textbackslash{}u2019ve just fallen in love with this advent calendar! \\ Awesome! I wanna one for my kids!\textbackslash{}r\textbackslash{}nRob: I used to get one every year as a \\ child! Loved them! \textbackslash{}r\textbackslash{}nEmma: Yeah, i remember! they were filled with \\ chocolates!\textbackslash{}r\textbackslash{}nLauren: they are different these days! much more \\ sophisticated! Haha!\textbackslash{}r\textbackslash{}nRob: yeah, they can be fabric/ wooden, shop bought/ \\ homemade, filled with various stuff\textbackslash{}r\textbackslash{}nEmma: what do you fit \\ inside?\textbackslash{}r\textbackslash{}nLauren: small toys, Christmas decorations, creative stuff, hair \\ bands \& clips, stickers, pencils \& rubbers, small puzzles, \\ sweets\textbackslash{}r\textbackslash{}nEmma: WOW! That\textbackslash{}u2019s brill! X\textbackslash{}r\textbackslash{}nLauren: i add one more very \\ special thing as well- little notes asking my children to do something nice \\ for someone else\textbackslash{}r\textbackslash{}nRob: i like that! My sister adds notes asking her kids \\ questions about christmas such as What did the 3 wise men bring? \\ etc\textbackslash{}r\textbackslash{}nLauren: i reckon it prepares them for Christmas \textbackslash{}r\textbackslash{}nEmma: and makes it \\ more about traditions and being kind to other people\textbackslash{}r\textbackslash{}nLauren: my children \\ get very excited every time they get one!\textbackslash{}r\textbackslash{}nEmma: i can see why! :)\end{tabular} \\ \midrule
Summary &
  \begin{tabular}[c]{@{}l@{}}1. Tomas Medina Caracas was a fugitive from a U.S. drug \\ trafficking indictment.\textbackslash{}n2. \textbackslash{}"El Negro Acacio\textbackslash{}" allegedly helped \\ manage extensive cocaine network.\textbackslash{}n3. U.S. Justice Department indicted him in \\ 2002.\textbackslash{}n4. Colombian military: He was killed in an attack on a guerrilla encampment.\end{tabular} &
  \begin{tabular}[c]{@{}l@{}}1. Emma and Rob love the advent calendar.\textbackslash{}n2. Lauren fits \\ inside calendar various items, for instance, small toys and Christmas \\ decorations.\textbackslash{}n3. Her children are excited whenever they get the calendar.\end{tabular} \\ \midrule
Wh-query &
  \begin{tabular}[c]{@{}l@{}}1. Who was Tomas Medina Caracas?\textbackslash{}n2. What was he indicted \\ for?\textbackslash{}n3. When was he indicted?\textbackslash{}n4. How did he die?\end{tabular} &
  \begin{tabular}[c]{@{}l@{}}1. What are Emma and Rob's attitude towards advent \\ calendar?\textbackslash{}n2. What does Lauren fit inside advent calendar?\textbackslash{}n3. What is the \\ reaction of Lauren's children when they get the calendar?\end{tabular} \\ \midrule
Yes/no Query &
  \begin{tabular}[c]{@{}l@{}}1. Yes: Was Tomas Medina Caracas a fugitive?\textbackslash{}n2. No: Did \\ \textbackslash{}"El Negro Acacio\textbackslash{}" help to fight against drug?\textbackslash{}n3. Yes: Was he \\ indicted by U.S. Justice Department?\textbackslash{}n4. No: Is he still alive?\end{tabular} &
  \begin{tabular}[c]{@{}l@{}}1. Yes: Do Emma and Rob love the advent calendar?\textbackslash{}n2. No: Is \\ Lauren unenthusiastic about advent calendar?\textbackslash{}n3. Yes: Do Lauren's children \\ enjoy receiving the calendar?\end{tabular} \\ \midrule
Wh-Instruction &
  \multicolumn{2}{l}{For each summary, write a general question about the article that can be answered by it} \\ \midrule
Yes/no Instruction &
  \multicolumn{2}{l}{For each summary, write a binary question about the article that can be answered by it} \\ \bottomrule
\end{tabular}%
}
\caption{One-shot prompt examples for news and dialogue domain.}
\label{tab:prompt_example}
\end{table*}
\label{sec:appendix}

% This is an appendix.

\end{document}